\documentclass{article}

\usepackage[preprint]{template/neurips_2024}

\usepackage{amsmath,amsfonts,bm}

\def\eqref#1{equation~\ref{#1}}

\def\1{\bm{1}}

\DeclareMathAlphabet{\mathsfit}{\encodingdefault}{\sfdefault}{m}{sl}
\SetMathAlphabet{\mathsfit}{bold}{\encodingdefault}{\sfdefault}{bx}{n}

\usepackage[table,xcdraw]{xcolor}
\usepackage{xurl}
\usepackage[hidelinks,breaklinks]{hyperref}
\usepackage{graphicx}
\usepackage{booktabs}
\usepackage{amsmath,amssymb}
\usepackage{afterpage}
\usepackage{caption}
\usepackage{multirow}
\usepackage{algorithm}
\usepackage{algpseudocode}
\usepackage{subcaption}
\usepackage{wrapfig}
\usepackage{amsthm}
\usepackage{array}
\usepackage{xspace}
\usepackage{placeins}
\usepackage{float}

\makeatletter
\renewcommand{\@noticestring}{Preprint.}
\makeatother

\usepackage{eso-pic}
\AddToShipoutPictureBG*{%
  \AtPageUpperLeft{%
    \raisebox{-2\baselineskip}{%
      \makebox[\paperwidth]{%
        \parbox{0.8\paperwidth}{\centering\footnotesize
          This work has been submitted to the IEEE for possible publication.
          Copyright may be transferred without notice, after which this
          version may no longer be accessible.}}}}}

\graphicspath{{figures/}}

\newcommand{\method}{\textsc{MixQuant}\xspace}
\newcommand{\updated}[1]{#1}

\let\cite\citep
\setcitestyle{authoryear,round,citesep={,},aysep={},yysep={;}}

\title{MixQuant: Adaptive Mixed-Precision Quantization for Large Language Models}

\author{%
  Ashitabh Misra\\
  University of Illinois at Urbana-Champaign\\
  \texttt{misra8@illinois.edu}
  \And
  Madhav Agrawal\\
  University of Illinois at Urbana-Champaign\\
  \texttt{madhav5@illinois.edu}
  \And
  Arham Jain\\
  University of Illinois at Urbana-Champaign\\
  \texttt{arhamj3@illinois.edu}
  \AND
  Tarek Abdelzaher\\
  University of Illinois at Urbana-Champaign\\
  \texttt{zaher@illinois.edu}
}

\begin{document}

\maketitle

\begin{abstract}
\updated{
Mixed-precision quantization improves the accuracy of post-training
quantization by allocating higher bitwidths to sensitive layers, but existing
methods solve the allocation for a single fixed memory budget. In practice
the budget varies across deployments and is unknown at calibration time.
Adaptive quantization addresses this with one offline calibration that serves
any budget, yet current methods score layer sensitivity in a manner that does not consider its dependency on quantization levels of other layers. We show that a layer's sensitivity depends strongly on the
bitwidths of its upstream layers and that this dependence shifts the
resulting preferred bit allocation. We propose \method{}, a technique-agnostic adaptive
framework that wraps any base quantizer. \method{} marginalizes each layer's
distortion over random quantized upstream configurations to obtain
budget-agnostic scores, calibrates the quantizer's parameters on plans the
allocator itself produces, and penalizes allocations that leave layers at the
lowest bitwidths. A single greedy pass then serves any budget at deployment.
Across Llama-3.2-3B, Llama-2-7B, and Mistral-7B under AWQ and GPTQ,
\method{} outperforms adaptive and mixed-precision baselines in every
setting, improving average accuracy by up to 8 points and reducing perplexity
from 12.43 to 10.70 at the tightest budget, while matching an ILP solver at
negligible deployment cost.}
\end{abstract}

\section{Introduction}
Large Language Models (LLMs) underpin a wide range of applications, from dialog
systems and code assistants to search and document understanding
\cite{llama2,rag,Islam2024-se}. Their
capability scales with size, and so does their cost: serving a model with tens of
billions of parameters exceeds the memory and bandwidth of most deployment
targets \cite{llama2,vllm}. Among compression techniques such as pruning, distillation, and
low-rank factorization \cite{liu2018rethinking,Hinton2015DistillingTK,parhi2023tensor}, post-training quantization (PTQ) is the most
widely adopted, reducing weight precision with only a small calibration set and
no retraining \cite{DBLP:journals/corr/abs-2103-13630,quantization_dnn_survey}.

Mixed-precision quantization extends uniform PTQ by exploiting the fact that
layers differ widely in sensitivity: allocating higher bitwidths to sensitive
layers and lower bitwidths elsewhere yields better accuracy at the same memory
footprint \cite{wang2019haq,multiquant}. 
We show that allocations that have similar memory consumption can differ in quality (Table~\ref{tab:bitwidth-vs-quality}). 
 Memory budget alone does not determine accuracy. Where the bits are placed is what matters.
\updated{Existing mixed-precision quantization methods search for an optimal allocation under a \emph{fixed} resource constraint \cite{wang2019haq,koryakovskiy2023one}. In practice the constraint is neither single nor known: the same model is deployed across many devices with different memory capacities, and new deployment targets appear after calibration is complete. Moreover, if calibration is slower than the memory availability changes, the allocation is always stale. Many budgets make per-budget calibration computationally infeasible; unknown budgets make it impossible. A solution must therefore calibrate once, producing metrics from which a bit allocation for any budget is derived cheaply at deployment~\cite{jin2020adabits, yu2019any}.}

\begin{figure}[htbp]
    \centering
    \begin{subfigure}[t]{\textwidth}
        \centering
        \includegraphics[width=\textwidth]{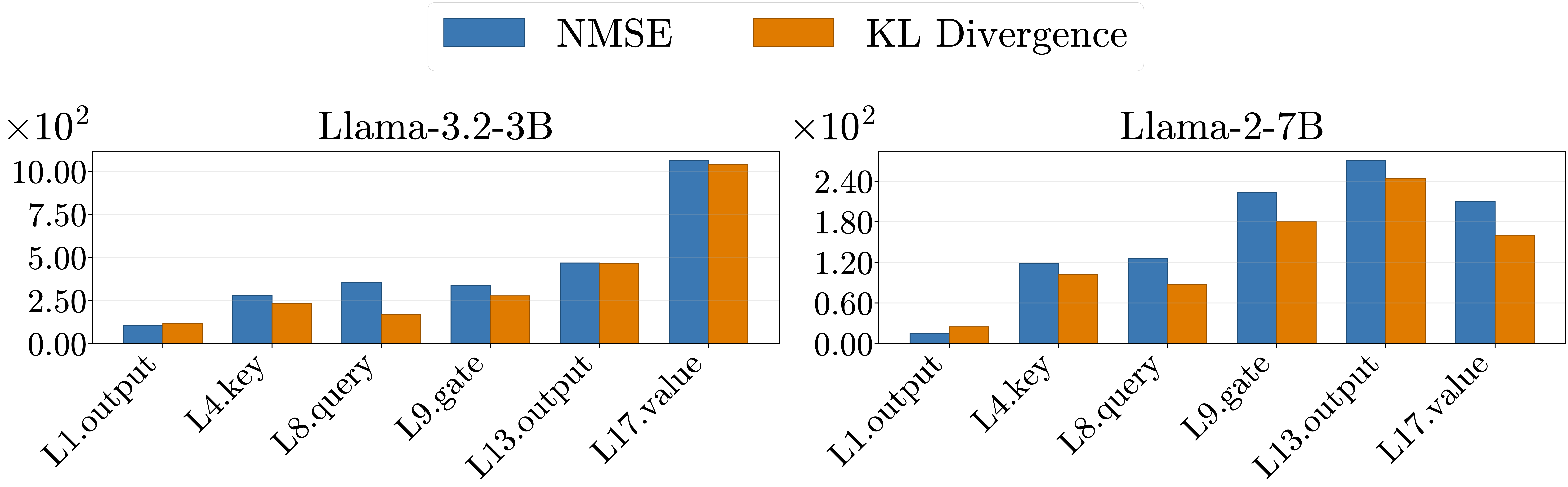}
        \caption{Metric variation across contexts. For six representative modules
        of Llama-3.2-3B (left) and Llama-2-7B (right), the relative spread
        $(\max-\min)/\min$ of NMSE and KL divergence across contexts. Both
        metrics move by up to two to three orders of magnitude.}
        \label{fig:metric-shift}
    \end{subfigure}\\[0.75em]
    \begin{subfigure}[t]{\textwidth}
        \centering
        \includegraphics[width=\textwidth]{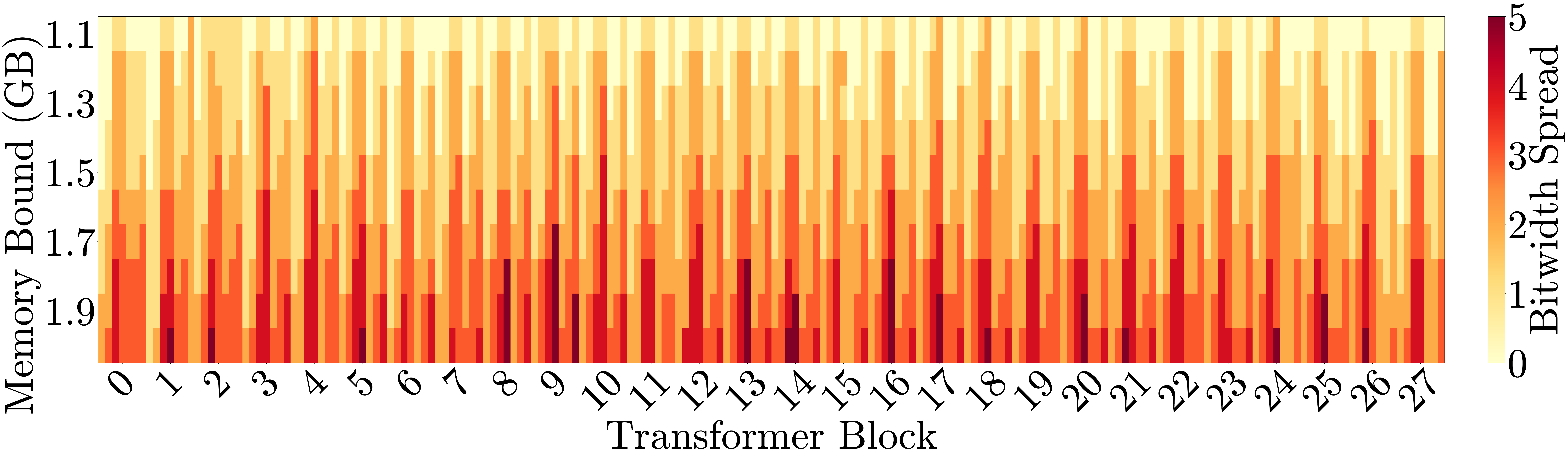}
        \caption{Allocation variation across contexts. Per-linear-layer bitwidth
        spread ($\max-\min$) for Llama-3.2-3B, one plan per context, from the ILP
        solver on the MCKP objective \eqref{eq:mckp}. The $x$-axis runs over
        linear layers, labelled by transformer block; the $y$-axis varies the
        memory bound. Spread appears across all memory bounds, showing that
        context choice shifts the bit allocation.}
        \label{fig:plan-spread}
    \end{subfigure}
    \caption{Effect of upstream context, measured over the same 20 random
    quantized contexts in both panels. Both per-module metrics (top) and the
    downstream plans they induce (bottom) vary with the context, so scoring in a
    single context biases the resulting bit allocation. \method{} removes this
    dependence by averaging over contexts (Section~\ref{sec:method-meanfield}).}
    \label{fig:context-distortion}
\end{figure}

Existing adaptive quantization methods, however, have three limitations. First, adaptive quantization techniques
developed for Convolutional Neural Networks rely on retraining to recover accuracy
\cite{jin2020adabits,bulat2021bit,robustquant,sun2024improved}, which is infeasible at LLM scale. 
\updated{Second, post-training adaptive quantization methods for LLMs, such as \cite{LIM}, compute their layer-wise metrics on inputs generated by upstream layers held in FP16 (we call the upstream bit allocation the \emph{context}). This context never occurs at deployment, where every layer is quantized, and we show that the choice of context shifts both the metrics and the resulting bit allocation (Figure~\ref{fig:context-distortion}).}
Third, methods such as Any-Precision LLM \cite{anyprecisionllm} are tied to a single underlying
quantization scheme, the non-uniform codebooks of SqueezeLLM \cite{squeezellm}, and do
not transfer across PTQ techniques.

\begin{figure}[t]
    \centering
    \includegraphics[width=\textwidth]{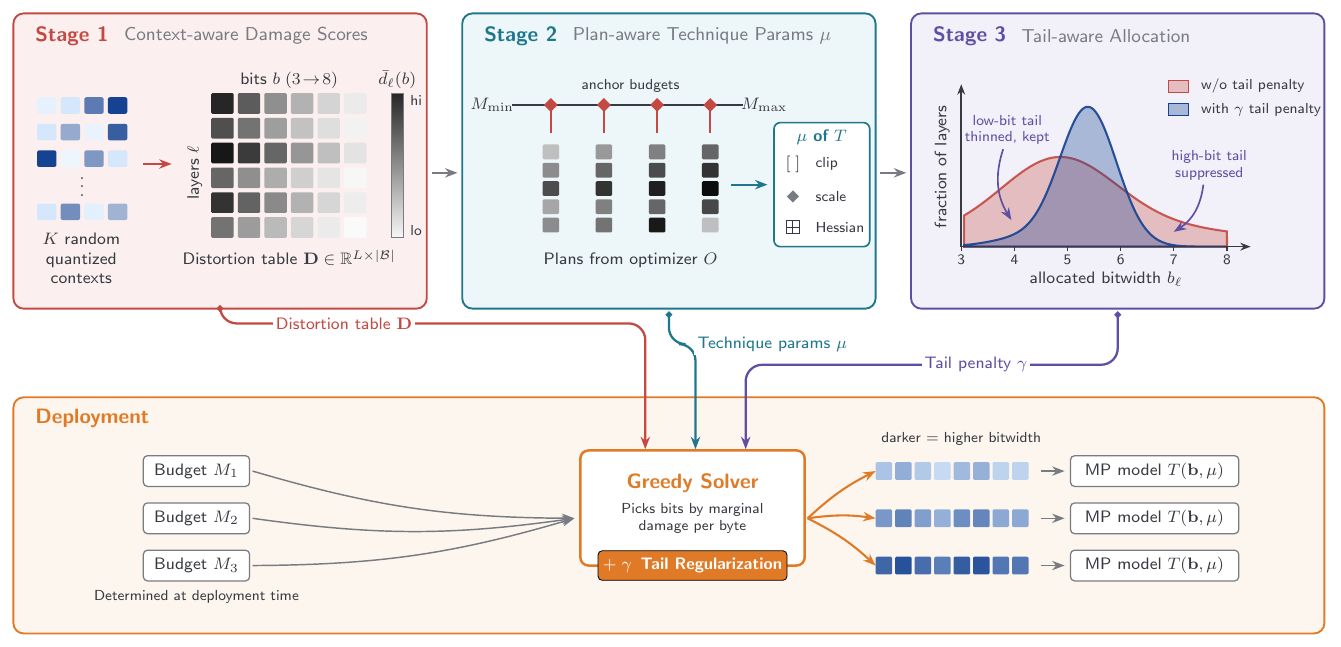}
    \caption{Overview of \method{}. \textbf{Stage 1} builds a budget-agnostic
distortion table $\mathbf{D}\in\mathbb{R}^{L\times|\mathcal{B}|}$ by scoring each
(module, bitwidth) pair under $K$ random quantized upstream contexts
(Section~\ref{sec:method-meanfield}). \textbf{Stage 2} estimates the technique parameters
$\mu$ on activations induced by anchor plans the greedy solver produces across
the feasible budget range, rather than on FP16 activations
(Section~\ref{sec:method-context}). \textbf{Stage 3} adds a tail penalty $\gamma$ that
steers spare budget away from the lowest bitwidths (Section~\ref{sec:tail-regularization}). At
deployment, each budget $M_i$ is served by a single greedy solve over
$\mathbf{D}$, $\mu$, and $\gamma$.}
    \label{fig:mq-method-overview}
\end{figure}

\updated{
\paragraph{Our Solution.} We propose \method{}, a technique-agnostic adaptive
PTQ framework that wraps any base quantizer, such as AWQ, GPTQ, and others \cite{awq,gptq}. It
addresses these limitations in three stages. 
\emph{First}, rather than scoring each layer against an FP16 network,
\method{} measures the output distortion of each (layer, bitwidth) pair under
random mixed-precision configurations of the layer's \emph{upstream} layers,
the only layers that shape its input. It then marginalizes this distortion
over the random configurations via Monte Carlo sampling.
Marginalization removes the dependence on the unknown
context, and the resulting distortion table depends only on the layer and its
bitwidth, so it serves every deployment budget. 
\emph{Second}, \method{}
re-introduces the information that is known at calibration time, namely the
technique and the solver. 
It estimates the technique parameters (AWQ scales
and clipping ranges, GPTQ Hessians) on activations induced by allocations the
solver actually produces, rather than on FP16 activations. 
\emph{Third},
because quantization error is irreversible along the forward pass, low-bit
assignments are disproportionately damaging; \method{} adds a \emph{tail
regularization} term that steers spare budget toward layers still at the
lowest bitwidths. 
Finally, we cast bit allocation as a multiple-choice
knapsack problem and solve it with a greedy algorithm cheap enough to run at
deployment time for each new budget. We empirically show that its
allocations match the downstream accuracy of an integer linear programming
solver (Table \ref{tab:mixquant-ablation}).
}

In summary, our main contributions are:
\begin{itemize}
    \item We identify deployment-context mismatch in adaptive quantization:
    per-layer sensitivity scores computed on the FP16 model do not reflect the
    fully quantized networks actually deployed. We propose a mean-field remedy,
    budget-agnostic distortion scores obtained by marginalizing each layer's
    error over random quantized configurations of its upstream layers.
    \item We develop \method{}, an end-to-end, technique-agnostic adaptive
    quantization pipeline that combines decoupled distortion scores, plan-aware
    estimation of technique parameters, and tail-regularized greedy allocation,
    so that a single offline calibration serves any memory budget at deployment
    with one cheap greedy solve.
    \item We evaluate \method{} across multiple models, datasets, and commonly
    used calibration metrics, and show that it consistently produces better bit
    allocations than existing adaptive and mixed-precision baselines
    \cite{LIM,coopq,hawqv2,jin2020adabits}.
\end{itemize}

\begin{table}[t]
    \centering
    \caption{Plans sampled from the main experiments
    (Tables~\ref{tab:llama-32-3b},~\ref{tab:llama-2-7b}, and~\ref{tab:results:mistral-llamastyle})
    at a fixed budget per model. Despite near-equal memory, neither perplexity
    nor accuracy tracks the unweighted average bitwidth: the highest-bitwidth
    plan is not the most accurate. How precision is distributed across layers,
    not the average, determines quality.}
    \label{tab:bitwidth-vs-quality}
    \small
    \begin{tabular}{@{}lcccc@{}}
    \toprule
    Model & Mem (GB) & Unweighted Avg Bitwidth & Wikitext PPL ($\downarrow$) & Avg Task Accuracy \\
    \midrule
    \multirow{6}{*}{Llama-3B} & \multirow{6}{*}{$\sim 1.25$}
          & 3.7          & 17.1            & 40.4          \\
        & & 3.9          & 16.6            & 44.2          \\
        & & 4.0          & \textbf{12.4}   & 47.8          \\
        & & 4.0          & 12.5            & \textbf{48.5} \\
        & & 4.3          & 12.9            & 46.9          \\
        & & \textbf{4.4} & 13.2            & 46.1          \\
    \midrule
    \multirow{6}{*}{Llama-7B} & \multirow{6}{*}{$\sim 3.0$}
          & 3.9          & 12.2            & 43.5          \\
        & & 4.0          & 11.9            & \textbf{45.0} \\
        & & 4.1          & \textbf{9.9}    & 44.3          \\
        & & 4.1          & 10.1            & 44.6          \\
        & & 4.3          & 10.3            & 42.9          \\
        & & \textbf{4.3} & 10.3            & 42.7          \\
    \bottomrule
    \end{tabular}
\end{table}

\section{Related Work}

\method{} builds on two lines of prior work: post-training quantization for LLMs, and adaptive quantization. We review each in turn.

\subsection{Post-Training Quantization for LLMs}
Since retraining billion-parameter models is prohibitively expensive,
\emph{Post-Training Quantization} (PTQ) has become the dominant approach
for compressing LLMs. Scaling-based methods redistribute quantization
difficulty between weights and activations: SmoothQuant \cite{smoothquant} migrates
activation outliers into the weights via a per-channel equivalent transform,
while AWQ \cite{awq} protects the few salient weight channels identified by
activation magnitude. GPTQ \cite{gptq} instead exploits second-order
information, quantizing column by column and using an approximate Hessian
to compensate the error introduced at each step. LQER \cite{lqer} takes a corrective route, absorbing the residual quantization error into an activation-scaled low-rank term. Rotation-based methods instead precondition the network: QuaRot \cite{quarot} applies
randomized Hadamard rotations and SpinQuant \cite{spinquant} learns them, both
suppressing outliers before quantization. \emph{Mixed-precision} methods instead allocate bits non-uniformly across the model: HAWQv2 \cite{hawqv2} assigns layer bit-widths by average Hessian trace, AMQ \cite{amq} searches per-layer precisions to minimize accuracy loss under a budget, and AQLM \cite{aqlm} pushes weights to $\sim$2 bits via
learned additive codebooks, all trading precision against a target budget. The aforementioned techniques assume the deployment memory budget is known a priori. In
real-time deployment, where constraints change at runtime, this assumption
fails---and re-solving the allocation for every new budget is impractical.

\subsection{Adaptive Quantization}
\emph{Adaptive quantization} removes this dependence: a single offline
calibration yields per-layer scores from which an allocation for \emph{any}
budget is recovered by one cheap solve at deployment. Early adaptive quantization methods, developed for CNNs and small transformers, achieve this through quantization-aware retraining: AdaBits~\cite{jin2020adabits}
jointly trains a single model to run at any of a fixed set of bit-widths,
BitMixer~\cite{bulat2021bit} trains a meta-network that selects arbitrary
mixed-precision configurations at runtime, and RobustQuant~\cite{robustquant}
regularizes weight kurtosis so layer distributions stay resilient across
precisions. All recover accuracy through gradient updates, which is infeasible
at LLM scale; we therefore focus on retraining-free adaptive quantization for LLMs. LIM~\cite{LIM}
scores each transformer block by the cosine similarity between its input and
output hidden states, a single-pass metric requiring no solver. CoopQ~\cite{coopq}
attributes a contribution to each layer via Shapley values and allocates bits
with an ILP. Any-Precision LLM~\cite{anyprecisionllm} stores one model that
serves multiple bit-widths, built on the non-uniform codebooks of
SqueezeLLM~\cite{squeezellm}.

\section{Problem Setup}
\label{sec:problem-setup}

\paragraph{Setup.} A pretrained LLM contains linear modules $\ell \in \{1,
\dots, L\}$, where module $\ell$ has $P_\ell$ parameters. Each module is
quantized by a PTQ technique $T$, e.g., AWQ or GPTQ \cite{awq,gptq}. 
The technique has
calibration parameters $\mu$, such as smoothing factors and clipping ranges
for AWQ or Hessian estimates for GPTQ. 
Bitwidths are chosen from an ordered
set $\mathcal{B}$ of admissible values ($\mathcal{B} = \{3, \dots, 8\}$ in our
experiments), and a \emph{plan} is a vector $\mathbf{b} = (b_1, \dots, b_L)
\in \mathcal{B}^L$. A plan occupies weight memory
$\mathrm{mem}(\mathbf{b}) = \tfrac{1}{8}\sum_{\ell=1}^{L} P_\ell\, b_\ell$
(reported in bytes).
Given a memory budget $M$, the deployment objective is
\begin{equation}
\min_{\mathbf{b},\, \mu} \;\; \mathcal{L}\big(T(\mathbf{b}, \mu)\big)
\quad \text{s.t.} \quad \mathrm{mem}(\mathbf{b}) \le M,
\label{eq:deployment-objective}
\end{equation}
where $\mathcal{L}$ is the loss of the quantized model. In the adaptive
setting, the technique $T$ and the allocation solver are fixed before
calibration, but the budget $M$ is a \emph{runtime} constraint unknown at calibration time.
\updated{This variation calls for a shared calibration and precomputed metrics independent of $M$.}

\paragraph{Surrogate.} \updated{Optimizing \eqref{eq:deployment-objective} directly is
intractable, since $\mathcal{L}$ is a black box over a configuration space
that is exponential in the number of modules. We replace $\mathcal{L}$ with a
sum of per-module distortion scores $\bar{d}_\ell(b)$, where $\bar{d}_\ell(b)$
measures the output distortion of quantizing module $\ell$ to $b$ bits. We
formulate the resulting allocation problem as a multiple-choice knapsack
problem (MCKP), a formulation shared by prior mixed-precision 
methods \cite{hawqv3,llm_mq,reactquant}}, 
\begin{equation}
\min_{x} \; \sum_{\ell=1}^{L} \sum_{b \in \mathcal{B}} \bar{d}_\ell(b)\,
x_{\ell,b}
\quad \text{s.t.} \quad
\sum_{b \in \mathcal{B}} x_{\ell,b} = 1 \;\; \forall \ell, \qquad
\frac{1}{8}\sum_{\ell=1}^{L} \sum_{b \in \mathcal{B}} P_\ell\, b\, x_{\ell,b} \le M,
\label{eq:mckp}
\end{equation}
\updated{where $x_{\ell,b} \in \{0,1\}$ indicates the assignment of bitwidth $b$ to
module $\ell$.
The additive objective presumes each module's score is independent of the
rest of the plan; Section~\ref{sec:method-meanfield} constructs
$\bar{d}_\ell(b)$ so that this holds by definition.
The technique parameters $\mu$ do not appear in \eqref{eq:mckp}; they are
estimated once during calibration and then held fixed
(Section~\ref{sec:method-context}).
}

\section{Method}
\label{sec:method}

\method{} solves the allocation problem
\eqref{eq:mckp} in three stages, which together estimate its two unknowns:
the distortion scores $\bar{d}_\ell(b)$ and the technique parameters $\mu$.
Figure~\ref{fig:mq-method-overview} gives an overview.
\textbf{Stage 1} estimates $\bar{d}_\ell(b)$. A module's distortion depends on
its \emph{context}, the bitwidths of the upstream modules that shape its
input, and the deployed context is unknown at calibration time. We therefore
average each score over random quantized upstream configurations, which yields
a budget-agnostic distortion table (Section~\ref{sec:method-meanfield}).
\textbf{Stage 2} estimates $\mu$. Unlike the budget, the technique and the
solver are known before deployment. We generate plans across the feasible
budget range with the greedy solver, and calibrate $\mu$ on the activations
these plans induce rather than on FP16 activations
(Section~\ref{sec:method-context}).
\textbf{Stage 3} introduces a tail regularizer. Distortion introduced by a
low-bit module cannot be undone by any downstream module, however high its
precision, and the per-module scores in \eqref{eq:mckp} do not capture this
cost. The regularizer biases the solver against leaving modules at the lowest
bitwidths (Section~\ref{sec:tail-regularization}).

\subsection{Stage 1: Mean-Field Distortion Estimation}
\label{sec:method-meanfield}

Any per-module distortion score is measured in a \emph{context}: the bitwidths
of the upstream modules that shape the input the module receives. Prior work
fixes this context to full precision. Layer- and block-wise PTQ reconstructs
each unit from FP16 inputs \cite{adaround,brecq}. Mixed-precision methods
rank the sensitivity of a layer while every other layer is left unquantized
\cite{hawqv2,hawqv3}. Activation-aware LLM quantizers calibrate their
transforms on FP16 activations \cite{awq,smoothquant}. A fully FP16 upstream
is a configuration that never occurs at deployment, where every module is
quantized. A representative score must instead be measured under quantized
upstream modules. 
The deployed plan is unknown at calibration time, so we construct a score table that does not depend on it and therefore serves every budget.

\paragraph{Context-dependent distortion.} The input to module $\ell$ drifts
from its FP16 value whenever upstream modules are quantized. A per-module
score therefore depends on the bitwidth of the module and on its upstream
context $\mathbf{b}_{<\ell} \in \mathcal{B}^{\ell-1}$. We instantiate
$d_\ell$ as the normalized mean squared error (NMSE) of the module output,
\begin{equation}
\label{eq:nmse}
d_\ell(b; \mathbf{b}_{<\ell})
= \frac{\mathbb{E}_x \big\| y_\ell^{(b,\,\mathbf{b}_{<\ell})}(x) -
y_\ell^{\mathrm{fp16}}(x) \big\|^2}
{\mathbb{E}_x \big\| y_\ell^{\mathrm{fp16}}(x) \big\|^2},
\end{equation}
where $x$ is a calibration sample, $y_\ell^{(b,\,\mathbf{b}_{<\ell})}$ is the
output of module $\ell$ quantized to $b$ bits under context
$\mathbf{b}_{<\ell}$, and $y_\ell^{\mathrm{fp16}}$ is the FP16 reference. The
normalization makes scores comparable across modules.

This dependence is large in practice, in both the raw metrics and the allocation they induce (Figure~\ref{fig:context-distortion}),
so a score measured in a single arbitrary context is not representative.

\paragraph{Averaging over quantized contexts.} The deployed context is
unknown at calibration time, so we treat it as random. We draw the upstream
bit-widths independently and uniformly over $\mathcal{B}$, written
$\pi(\mathbf{b}_{<\ell}) = \prod_{m<\ell} \mathrm{Unif}(\mathcal{B})$, and
define the \emph{decoupled distortion} $\bar{d}_\ell(b)$ as the expected NMSE
under this prior. We estimate it with $K$ Monte Carlo draws,
\begin{equation}
\label{eq:marginal}
\bar{d}_\ell(b) =
\mathbb{E}_{\mathbf{B}_{<\ell}\sim\pi}\!\big[d_\ell(b;
\mathbf{B}_{<\ell})\big],
\qquad
\hat{d}_\ell(b) = \frac{1}{K}\sum_{k=1}^{K} d_\ell\big(b;
\mathbf{b}_{<\ell}^{(k)}\big),
\quad \mathbf{b}_{<\ell}^{(k)} \sim \pi.
\end{equation}

For the sampled context $\mathbf{b}_{<\ell}^{(k)}$, the technique parameters ($\mu$) of the quantized modules are recalibrated under that context, so no measurement in the table
depends on FP16 calibration.
These per-context parameters are used only for scoring and are then discarded. The single set of parameters deployed with the model is estimated in Stage 2.

The resulting table depends on $(\ell, b)$ alone and serves every budget. 
This averaging is a mean-field decoupling: the unknown joint configuration of upstream modules is replaced by its average effect under the fixed prior $\pi$, so each
module's score becomes independent of the decisions made for other modules.

\subsection{Stage 2: Plan-Aware Technique Parameters}
\label{sec:method-context}

\updated{Stage 1 removes all plan information from the scores, but not all of it is
unknown. The technique $T$ and the solver are fixed before deployment, and
the solver has its own systematic preferences over plans. We re-introduce
this known structure through the technique parameters $\mu$, by calibrating
them on plans the solver itself produces.}

\updated{We partition the feasible memory range $[M_{\min}, M_{\max}]$ into $I$
intervals, where $M_{\min}$ places every module at the smallest bitwidth and
$M_{\max}$ at the largest. 
From interval $i$ we sample $J$ budgets and solve each with the same allocator used at deployment (Section~\ref{sec:greedy-allocator}), which yields $J$ anchor plans.
We quantize the network
according to each anchor plan, calibrate one parameter set on the activations
it induces, and average the $J$ sets to form $\mu_i$.}

\updated{For GPTQ the parameters are the layer Hessians, for AWQ the smoothing factors
and clipping ranges. Averaging is exact for Hessians, since they are second
moments of the inputs and their mean equals the Hessian of the mixed
activations. For AWQ it is a heuristic, and we use it uniformly across
techniques as a simple way to capture the average behavior the technique and
solver induce. Bucketing exists because plans at opposite ends of the budget
range induce different activation statistics, so one parameter set fits
neither end well, while within an interval the anchor plans are close enough
to share one. At deployment, a budget $M$ falling in interval $i$ is served
by one greedy solve and the precomputed $\mu_i$.}

\subsection{Stage 3: Tail Regularization}
\label{sec:tail-regularization}

Quantization distortion is irreversible along the forward pass. Once a
module's output is corrupted by a low-bit assignment, no downstream module
can recover the lost information, however high its precision, and the error
compounds as it propagates. The lowest bitwidths should therefore be conceded
sparingly. We make this explicit with a penalty that steers spare budget
toward the modules still at the lowest bitwidths, so the lowest precisions
are used only when the budget leaves no alternative.

Let $b_0$ be the smallest bit-width in $\mathcal{B}$, and let $b^{+}$ denote the successor of $b$ in the ordered set
$\mathcal{B}$. Raising a module from $b$ to $b^{+}$ removes distortion at the
cost of additional memory. We score this upgrade by the distortion removed per
additional memory,
\begin{equation}
e_\ell(b) \;=\; \frac{\bar{d}_\ell(b) - \bar{d}_\ell(b^{+})}{P_\ell\,(b^{+} - b)},
\label{eq:efficiency}
\end{equation}
and discount it by how far the module has already been raised above the
floor,
\begin{equation}
\tilde{e}_\ell(b) \;=\; \frac{e_\ell(b)}{1 + \gamma\,(b - b_0)},
\qquad \gamma \ge 0.
\label{eq:tail-penalty}
\end{equation}
The depth $b - b_0$ measures how far into the tail an upgrade reaches. Each
additional bit granted to an already-raised module is progressively
discounted relative to a module still at the floor, so the budget is pulled
toward clearing modules off the lowest bitwidths rather than enriching a few
modules deeply. The strength $\gamma$ controls how strongly low bitwidths are
discouraged.

\subsection{Greedy Allocation}
\label{sec:greedy-allocator}

Given a feasible budget $M$, the allocator starts from the uniform floor plan
$\mathbf{b} = (b_0, \dots, b_0)$ and maintains a set of candidate upgrades,
one per module: the transition from its current bitwidth $b_\ell$ to the
successor $b_\ell^{+}$, scored by the penalized efficiency
$\tilde{e}_\ell(b_\ell)$ of \eqref{eq:tail-penalty}. At each step the
highest-scoring upgrade that fits the remaining budget is applied. The
upgraded module's old transition is removed and its next transition, from
$b_\ell^{+}$ onward, is inserted. The procedure ends when no candidate fits
the remaining budget. Algorithm~\ref{alg:greedy} gives the pseudocode. The
solve performs at most $L(|\mathcal{B}|-1)$ upgrades, each selecting the best
of at most $L$ candidates, so a full allocation costs
$O(L\,|\mathcal{B}| \log L)$ with a priority queue, and is negligible at
deployment time.

\begin{algorithm}[t]
\caption{Tail-Regularized Greedy Allocation}
\label{alg:greedy}
\begin{algorithmic}[1]
\Require distortion table $\bar{d}$, budget $M$, penalty $\gamma$
\State $b_0 \gets$ smallest uniform bitwidth in $\mathcal{B}$
\State $b_\ell \gets b_0$ for all $\ell$;\quad $R \gets M - \mathrm{mem}(\mathbf{b})$ \Comment{remaining budget}
\State $Q \gets \{(\tilde{e}_\ell(b_\ell),\, \ell) : b_\ell < b_{\max}\}$ \Comment{candidate transitions}
\While{$Q$ contains a transition with cost $P_\ell\,(b_\ell^{+} - b_\ell) \le R$}
    \State pop the highest-scoring $(\tilde{e}_\ell,\, \ell)$ with $P_\ell\,(b_\ell^{+} - b_\ell) \le R$
    \State $R \gets R - P_\ell\,(b_\ell^{+} - b_\ell)$;\quad $b_\ell \gets b_\ell^{+}$
    \If{$b_\ell < b_{\max}$}
        \State push $(\tilde{e}_\ell(b_\ell),\, \ell)$ into $Q$
    \EndIf
\EndWhile
\State \Return $\mathbf{b}$
\end{algorithmic}
\end{algorithm}
\section{Evaluation}
\label{sec:results}

\paragraph{Setup.} We evaluate \method{} on Llama-3.2-3B, Llama-2-7B, and
Mistral-7B-v0.1. Plans assign per-linear-module bitwidths from
$\mathcal{B} = \{3, \dots, 8\}$ using AWQ or GPTQ as the base technique, with
8-bit activations. Quantization is simulated (quantize--dequantize to FP16);
we report memory as packed linear-weight size, excluding scale overhead.
Distortion tables and technique parameters are calibrated on 128 random
windows of length 2048 from the WikiText-2 training split. Unless stated
otherwise, \method{} uses $K = 10$ Monte Carlo context draws per
(module, bitwidth) pair (Section~\ref{sec:method-meanfield}), $I = 3$ budget
intervals with $J = 5$ anchor plans each (Section~\ref{sec:method-context}),
and tail penalty $\gamma = 10$ (Section~\ref{sec:tail-regularization}).
We evaluate word-level perplexity on the WikiText-2 validation split and
accuracy on PIQA, ARC-Challenge, WinoGrande, and MMLU (grouped into
humanities, social sciences, STEM, and other), all through the EleutherAI
lm-evaluation-harness \cite{eval-harness}.

\paragraph{Baselines.} The baselines span the
space of FP16-prior scoring metrics, alongside existing adaptive pipelines.
The first four are controlled comparisons constructed by us. Each scores
every (module, bitwidth) pair against the FP16 model and feeds the resulting
table to the same ILP solver, based on the MCKP formulation of
\eqref{eq:mckp}. The four therefore differ only in the scoring metric.
\emph{Quant.\ Err.} and \emph{KL Div} compare the output activations of each
quantized module to their full-precision counterparts, scoring the pair by
the activation error and the KL divergence of the output distributions,
respectively. \emph{HAWQ-v2} adopts the metric of HAWQ-V2 \cite{hawqv2}, which
combines a Hessian-based importance estimate with the quantization error, so
that both the sensitivity of a module and the magnitude of its perturbation
enter the score. \emph{Fisher} uses the Fisher approximation of the Hessian
as a pure importance estimate, capturing sensitivity alone. Together these
four cover error, distributional divergence, importance, and their
combination, the principal axes along which prior scoring metrics differ.

The remaining two baselines are existing adaptive quantization methods and
run with their own allocation procedures. \emph{LIM} \cite{LIM} scores modules
by the cosine similarity between input and output of each transformer block. By design, 
LIM's methodology does not extend to linear-layer level granularity.
\emph{CoopQ} \cite{coopq} departs from per-module scoring altogether, estimating
module contributions via Shapley-value distributions and solving a novel ILP
formulation for adaptive quantization.

\paragraph{Accuracy across memory budgets.}
Tables~\ref{tab:llama-32-3b},~\ref{tab:llama-2-7b}, and~\ref{tab:results:mistral-llamastyle} report perplexity and
downstream accuracy under three memory budgets per model. \method{} attains
the best average accuracy and the lowest perplexity in every
(model, budget, technique) cell, for both AWQ and GPTQ, which supports the
claim that the pipeline is agnostic to the base technique.

\begin{table}[t]
\centering
\caption{Downstream mixed-precision quantization results on Llama-3.2-3B.
 Accuracies (\%); Wiki.\ = word perplexity ($\downarrow$).
  Avg excludes Wiki. \textbf{Bold} = best, \underline{underline} = second best per column within each memory budget.}
\label{tab:llama-32-3b}
\renewcommand{\arraystretch}{1.3}
\setlength{\tabcolsep}{4pt}
\resizebox{\textwidth}{!}{%
\begin{tabular}{c c *{8}{c}!{\vrule width 0.6pt}c!{\vrule width 0.6pt} *{8}{c}!{\vrule width 0.6pt}c}
\hline
 &  & \multicolumn{9}{c}{AWQ} & \multicolumn{9}{c}{GPTQ} \\ \cline{3-20}
Mem (GB) & Method & Wiki. & Arc & PIQA & Wino & STEM & Hum. & Soc. & Othr & Avg & Wiki. & Arc & PIQA & Wino & STEM & Hum. & Soc. & Othr & Avg \\
\hline
\multirow{7}{*}{1.25} & Quant. Err. & 13.26 & 34.3 & 71.3 & 62.1 & 33.8 & 34.7 & 43.0 & 43.1 & 46.1 & 15.75 & 32.8 & 70.2 & 61.6 & 31.7 & 30.5 & 36.0 & 35.4 & 42.6 \\
 & KL Div & 12.92 & 34.8 & 71.7 & 62.8 & 33.5 & 36.0 & 43.2 & 46.6 & 46.9 & 15.39 & 33.6 & 70.0 & 61.1 & 34.0 & 31.5 & 37.6 & 37.2 & 43.6 \\
 & HAWQ-v2 & \underline{12.43} & 34.4 & \underline{73.3} & \underline{64.2} & 35.1 & 37.0 & 44.6 & 46.0 & 47.8 & \underline{13.35} & \underline{35.8} & \underline{72.6} & \underline{63.9} & \underline{35.1} & \underline{34.7} & \underline{40.7} & \underline{39.5} & 46.0 \\
 & Fisher & 12.55 & \underline{35.6} & 73.1 & 63.9 & \underline{35.2} & \underline{37.7} & \underline{46.5} & \underline{47.6} & \underline{48.5} & 15.15 & 33.9 & 71.3 & 63.5 & 32.8 & 31.4 & 35.3 & 37.2 & 43.6 \\
 & CoopQ & 16.58 & 34.5 & 73.1 & 61.6 & 32.1 & 32.4 & 37.0 & 38.6 & 44.2 & 18.22 & 32.3 & 69.3 & 58.9 & 30.7 & 30.9 & 35.8 & 33.5 & 41.6 \\
 & LIM & 17.08 & 35.4 & 72.3 & 59.9 & 25.8 & 28.4 & 30.0 & 31.0 & 40.4 & 18.98 & 32.1 & 67.9 & 58.5 & 29.5 & 29.5 & 31.2 & 30.0 & 39.8 \\
\rowcolor{gray!15} & \textbf{\method{}} & \textbf{10.70} & \textbf{43.4} & \textbf{75.6} & \textbf{69.0} & \textbf{39.6} & \textbf{44.9} & \textbf{57.7} & \textbf{58.7} & \textbf{55.6} & \textbf{11.19} & \textbf{42.0} & \textbf{76.3} & \textbf{64.5} & \textbf{40.5} & \textbf{43.9} & \textbf{55.4} & \textbf{55.6} & \textbf{54.0} \\ \cline{1-20}
\multirow{7}{*}{1.5} & Quant. Err. & 11.02 & 39.2 & 73.9 & \underline{65.5} & 36.2 & 40.0 & 48.8 & 51.8 & 50.8 & \underline{10.81} & 39.2 & \underline{74.6} & 64.5 & \underline{37.6} & \underline{40.1} & \underline{47.5} & \underline{49.9} & 50.5 \\
 & KL Div & 11.09 & \underline{39.4} & 74.0 & 64.6 & 37.1 & 39.9 & 48.6 & 51.1 & 50.7 & 12.31 & 36.8 & 73.2 & \underline{64.9} & 35.0 & 36.3 & 46.0 & 45.7 & 48.3 \\
 & HAWQ-v2 & \underline{10.70} & 39.2 & \underline{75.0} & 63.5 & \underline{38.4} & \underline{41.5} & \underline{52.4} & \underline{53.4} & \underline{51.9} & 10.84 & \underline{39.5} & 74.3 & 64.2 & 36.0 & 37.1 & 44.8 & 47.5 & 49.1 \\
 & Fisher & 11.00 & 39.1 & 74.1 & 64.6 & 35.9 & 40.3 & 49.5 & 51.6 & 50.7 & 12.49 & 36.4 & 73.6 & 63.5 & 36.9 & 36.3 & 43.6 & 45.0 & 47.9 \\
 & CoopQ & 14.70 & 36.6 & 74.1 & 64.7 & 37.4 & 39.1 & 48.1 & 50.1 & 50.0 & 15.37 & 35.6 & 72.1 & 62.9 & 35.7 & 37.5 & 46.1 & 48.6 & 48.3 \\
 & LIM & 14.66 & 37.4 & 73.0 & 64.8 & 37.3 & 37.4 & 47.4 & 49.3 & 49.5 & 15.78 & 33.6 & 72.1 & 62.2 & 35.9 & 37.4 & 44.5 & 43.9 & 47.1 \\
\rowcolor{gray!15} & \textbf{\method{}} & \textbf{9.69} & \textbf{45.9} & \textbf{76.9} & \textbf{68.7} & \textbf{44.8} & \textbf{48.6} & \textbf{61.1} & \textbf{62.0} & \textbf{58.3} & \textbf{9.99} & \textbf{44.9} & \textbf{76.7} & \textbf{68.9} & \textbf{45.1} & \textbf{48.0} & \textbf{62.0} & \textbf{61.6} & \textbf{58.2} \\ \cline{1-20}
\multirow{7}{*}{1.75} & Quant. Err. & 9.87 & 43.6 & 76.6 & 66.6 & 42.6 & \underline{47.6} & \underline{60.0} & \underline{59.9} & \underline{56.7} & 9.90 & 43.5 & \underline{76.4} & 66.9 & 42.2 & \underline{47.1} & 59.0 & 56.9 & 56.0 \\
 & KL Div & 9.90 & 43.1 & 76.6 & 67.9 & 40.6 & 45.7 & 59.3 & 59.3 & 56.1 & 10.97 & 42.4 & 75.4 & 65.6 & 40.7 & 42.7 & 54.4 & 52.5 & 53.4 \\
 & HAWQ-v2 & \underline{9.85} & \underline{44.1} & \textbf{76.8} & 67.3 & \underline{43.0} & 47.0 & 59.5 & 59.3 & \underline{56.7} & \underline{9.87} & \underline{44.0} & \underline{76.4} & \underline{67.1} & 40.8 & 46.1 & 55.8 & 56.3 & 55.2 \\
 & Fisher & 10.02 & 41.9 & 76.2 & 68.0 & 39.5 & 45.2 & 58.3 & 57.2 & 55.2 & 11.20 & 40.4 & 75.3 & \underline{67.1} & 39.3 & 41.9 & 53.0 & 50.9 & 52.6 \\
 & CoopQ & 12.49 & 41.3 & 75.4 & \underline{68.3} & 42.4 & 44.6 & 59.0 & 57.2 & 55.4 & 13.00 & 38.8 & 74.3 & 64.9 & 43.0 & 45.2 & 56.2 & 55.4 & 54.0 \\
 & LIM & 12.48 & 39.6 & 74.6 & 67.4 & 42.9 & 45.5 & 57.8 & 58.9 & 55.3 & 13.35 & 39.6 & 74.2 & 65.7 & \underline{44.1} & 46.5 & \underline{60.1} & \underline{59.2} & 55.6 \\
\rowcolor{gray!15} & \textbf{\method{}} & \textbf{9.44} & \textbf{46.2} & \underline{76.7} & \textbf{69.1} & \textbf{45.9} & \textbf{49.3} & \textbf{63.6} & \textbf{62.6} & \textbf{59.0} & \textbf{9.51} & \textbf{46.3} & \textbf{77.0} & \textbf{69.1} & \textbf{45.5} & \textbf{49.1} & \textbf{62.7} & \textbf{62.5} & \textbf{58.9} \\ \cline{1-20}
\hline
\end{tabular}%
}
\end{table}

\begin{table}[t]
\centering
\caption{Downstream mixed-precision quantization results on Llama-2-7B. Accuracies (\%); Wiki.\ = word perplexity ($\downarrow$). 
Avg excludes Wiki. \textbf{Bold} = best, \underline{underline} = second best per column within each memory budget.}
\label{tab:llama-2-7b}
\renewcommand{\arraystretch}{1.3}
\setlength{\tabcolsep}{4pt}
\resizebox{\textwidth}{!}{%
\begin{tabular}{c c *{8}{c}!{\vrule width 0.6pt}c!{\vrule width 0.6pt} *{8}{c}!{\vrule width 0.6pt}c}
\hline
 &  & \multicolumn{9}{c}{AWQ} & \multicolumn{9}{c}{GPTQ} \\ \cline{3-20}
Mem (GB) & Method & Wiki. & Arc & PIQA & Wino & STEM & Hum. & Soc. & Othr & Avg & Wiki. & Arc & PIQA & Wino & STEM & Hum. & Soc. & Othr & Avg \\
\hline
\multirow{7}{*}{3.0} & Quant. Err. & 10.32 & 42.8 & 75.6 & 66.0 & 24.7 & 27.7 & 31.2 & 32.0 & 42.9 & 10.46 & 41.7 & 76.3 & 65.1 & 29.5 & 29.7 & 36.9 & 35.5 & 45.0 \\
 & KL Div & 10.36 & 42.3 & \underline{76.2} & 66.5 & 24.9 & 28.1 & 29.8 & 31.4 & 42.7 & 11.01 & 41.4 & 76.0 & 65.7 & 29.3 & 29.4 & 35.1 & 34.8 & 44.5 \\
 & HAWQ-v2 & \underline{9.98} & \underline{43.3} & 75.9 & \underline{67.4} & 26.4 & 30.1 & 32.9 & 33.9 & 44.3 & \underline{10.25} & \underline{42.5} & \underline{76.7} & 65.8 & 29.8 & \underline{31.2} & \underline{37.8} & 35.8 & 45.7 \\
 & Fisher & 10.12 & 43.0 & 76.0 & 67.1 & 27.6 & \underline{30.3} & 33.5 & 34.8 & 44.6 & 10.79 & 42.3 & \underline{76.7} & 65.2 & \underline{30.5} & 29.8 & 37.6 & \underline{36.0} & 45.4 \\
 & CoopQ & 11.94 & 41.2 & 75.9 & 67.2 & \underline{29.8} & 28.8 & \underline{35.1} & \underline{37.0} & \underline{45.0} & 11.96 & 39.8 & 76.2 & \underline{66.9} & 27.5 & \underline{31.2} & 30.9 & 32.5 & 43.6 \\
 & LIM & 12.20 & 40.9 & 75.9 & 66.1 & 29.2 & 28.8 & 30.9 & 32.9 & 43.5 & 12.10 & 40.9 & 75.5 & 66.5 & 24.6 & 30.4 & 28.5 & 31.1 & 42.5 \\
\rowcolor{gray!15} & \textbf{\method{}} & \textbf{9.50} & \textbf{45.2} & \textbf{77.3} & \textbf{69.1} & \textbf{31.3} & \textbf{35.9} & \textbf{40.9} & \textbf{41.8} & \textbf{48.8} & \textbf{9.70} & \textbf{44.9} & \textbf{77.5} & \textbf{69.1} & \textbf{31.8} & \textbf{35.7} & \textbf{40.2} & \textbf{39.7} & \textbf{48.4} \\ \cline{1-20}
\multirow{7}{*}{3.5} & Quant. Err. & 9.45 & 43.2 & 76.6 & \underline{69.1} & 30.8 & 35.2 & \underline{40.6} & 42.0 & \underline{48.2} & 9.57 & 42.6 & 77.3 & 65.9 & 32.3 & 32.9 & 40.8 & 39.1 & 47.3 \\
 & KL Div & 9.57 & 42.6 & 76.9 & 67.7 & 30.4 & 35.0 & 39.9 & 41.8 & 47.8 & 10.07 & 42.9 & 76.6 & 66.0 & \underline{33.2} & 31.2 & 39.7 & 36.6 & 46.6 \\
 & HAWQ-v2 & \underline{9.42} & 43.9 & 76.6 & 67.8 & 29.8 & \underline{35.7} & 40.4 & \underline{42.4} & 48.1 & \underline{9.53} & \underline{43.0} & \underline{77.5} & 66.2 & 33.1 & 33.7 & \underline{41.1} & 39.6 & 47.7 \\
 & Fisher & 9.51 & \underline{44.5} & 76.3 & \textbf{69.4} & 28.6 & 34.1 & 38.8 & 40.0 & 47.4 & 10.02 & 41.9 & 76.4 & 67.9 & 33.0 & 31.9 & 40.2 & 38.4 & 47.1 \\
 & CoopQ & 11.32 & 42.4 & \underline{77.3} & 67.6 & 28.2 & 29.7 & 30.8 & 35.1 & 44.5 & 11.25 & 42.2 & 77.3 & \textbf{68.7} & 30.0 & \underline{35.3} & 36.1 & \underline{39.7} & 47.0 \\
 & LIM & 11.48 & 43.1 & 76.6 & 66.5 & \underline{31.5} & 31.8 & 36.6 & 38.1 & 46.3 & 11.37 & 42.0 & 76.6 & \underline{68.5} & 28.0 & 33.8 & 34.7 & 36.0 & 45.7 \\
\rowcolor{gray!15} & \textbf{\method{}} & \textbf{9.06} & \textbf{45.8} & \textbf{77.7} & \underline{69.1} & \textbf{35.6} & \textbf{39.0} & \textbf{47.8} & \textbf{47.6} & \textbf{51.8} & \textbf{9.12} & \textbf{46.1} & \textbf{78.0} & \textbf{68.7} & \textbf{33.7} & \textbf{37.9} & \textbf{45.7} & \textbf{45.2} & \textbf{50.8} \\ \cline{1-20}
\multirow{7}{*}{4.0} & Quant. Err. & 9.12 & 45.5 & \underline{77.5} & 68.4 & 33.7 & 38.6 & 45.1 & 45.5 & \underline{50.6} & \underline{9.13} & 44.5 & 77.5 & 66.9 & \textbf{35.4} & 38.2 & \textbf{47.4} & \underline{46.5} & 50.9 \\
 & KL Div & 9.17 & \textbf{46.6} & 77.3 & 68.4 & 34.2 & 37.4 & 44.8 & 44.6 & 50.5 & 9.42 & \textbf{46.1} & 77.5 & 67.1 & 34.5 & \underline{38.3} & 45.9 & 45.8 & 50.7 \\
 & HAWQ-v2 & \underline{9.09} & 45.2 & 77.2 & 68.5 & 33.1 & \underline{38.7} & \underline{45.6} & \underline{45.7} & \underline{50.6} & 9.14 & 45.4 & \underline{77.7} & 67.3 & \underline{35.0} & 38.0 & 46.5 & 45.4 & 50.8 \\
 & Fisher & 9.22 & 45.1 & 77.3 & 68.7 & 32.4 & 37.5 & 44.6 & 44.2 & 50.0 & 9.67 & 42.7 & 77.4 & \underline{68.3} & 33.9 & 33.9 & 43.1 & 41.2 & 48.6 \\
 & CoopQ & 10.68 & 45.1 & 77.0 & \underline{69.0} & 28.7 & 32.0 & 35.5 & 38.0 & 46.5 & 10.67 & 43.7 & 77.5 & 68.2 & 31.3 & 36.6 & 41.7 & 42.4 & 48.8 \\
 & LIM & 10.85 & 43.4 & \underline{77.5} & 68.4 & \underline{34.3} & 36.6 & 44.4 & 45.6 & 50.0 & 10.75 & 44.2 & 77.1 & \textbf{68.8} & 30.2 & 36.3 & 39.7 & 42.0 & 48.3 \\
\rowcolor{gray!15} & \textbf{\method{}} & \textbf{8.95} & \underline{46.2} & \textbf{77.8} & \textbf{69.1} & \textbf{35.1} & \textbf{39.7} & \textbf{47.8} & \textbf{48.8} & \textbf{52.1} & \textbf{8.97} & \underline{45.8} & \textbf{78.2} & \textbf{68.8} & 34.9 & \textbf{40.2} & \underline{47.1} & \textbf{47.3} & \textbf{51.7} \\ \cline{1-20}
\hline
\end{tabular}%
}
\end{table}

\begin{table}[t]
\centering
\caption{Downstream mixed-precision quantization results on Mistral-7B.
 Accuracies (\%); Wiki.\ = word perplexity ($\downarrow$).
  Avg excludes Wiki. \textbf{Bold} = best, \underline{underline} = second best per column within each memory budget.}
\label{tab:results:mistral-llamastyle}
\renewcommand{\arraystretch}{1.3}
\setlength{\tabcolsep}{4pt}
\resizebox{\textwidth}{!}{%
\begin{tabular}{c c *{8}{c}!{\vrule width 0.6pt}c!{\vrule width 0.6pt} *{8}{c}!{\vrule width 0.6pt}c}
\hline
 &  & \multicolumn{9}{c}{AWQ} & \multicolumn{9}{c}{GPTQ} \\ \cline{3-20}
Mem (GB) & Method & Wiki. & Arc & PIQA & Wino & STEM & Hum. & Soc. & Othr & Avg & Wiki. & Arc & PIQA & Wino & STEM & Hum. & Soc. & Othr & Avg \\
\hline
\multirow{7}{*}{3.0} & Quant. Err. & 9.36 & \underline{46.2} & 79.3 & 70.0 & \underline{43.9} & \underline{48.9} & \underline{63.3} & \underline{60.6} & \underline{58.9} & 9.51 & \underline{46.6} & \underline{79.0} & 70.8 & 43.5 & 45.3 & 58.1 & 56.1 & 57.1 \\
 & KL Div & \underline{9.31} & 46.0 & 79.4 & \underline{71.0} & 42.8 & 47.7 & 61.7 & 58.8 & 58.2 & \underline{9.42} & 46.2 & 78.7 & \textbf{71.7} & \underline{43.8} & 46.7 & 59.2 & 58.2 & 57.8 \\
 & HAWQ-v2 & 9.63 & 45.0 & \underline{79.5} & \underline{71.0} & 43.6 & 48.4 & 61.7 & 59.1 & 58.3 & 9.53 & 44.9 & 78.9 & 70.1 & 43.5 & \underline{46.8} & \underline{60.6} & \underline{58.6} & 57.6 \\
 & Fisher & 9.67 & 44.5 & 79.4 & 70.4 & 43.4 & 48.4 & 61.4 & 58.4 & 58.0 & 9.90 & 45.8 & 78.6 & 70.0 & 41.0 & 43.3 & 54.0 & 53.2 & 55.1 \\
 & CoopQ & 10.48 & 42.9 & 78.0 & 68.8 & 40.6 & 44.4 & 56.8 & 55.0 & 55.2 & 10.91 & 43.9 & 78.0 & 70.1 & 37.3 & 39.6 & 49.9 & 48.5 & 52.5 \\
 & LIM & 10.62 & 43.9 & 77.9 & 68.0 & 40.5 & 45.2 & 57.2 & 55.4 & 55.4 & 10.86 & 44.5 & 77.7 & 70.4 & 38.9 & 41.7 & 52.7 & 50.3 & 53.8 \\
\rowcolor{gray!15} & \textbf{\method{}} & \textbf{8.92} & \textbf{51.0} & \textbf{79.7} & \textbf{73.6} & \textbf{47.1} & \textbf{50.4} & \textbf{66.1} & \textbf{64.2} & \textbf{61.7} & \textbf{9.10} & \textbf{50.1} & \textbf{80.3} & \underline{71.6} & \textbf{47.0} & \textbf{50.8} & \textbf{66.7} & \textbf{64.2} & \textbf{61.5} \\ \cline{1-20}
\multirow{7}{*}{3.5} & Quant. Err. & 8.93 & 49.1 & 79.8 & 72.5 & 45.2 & 49.8 & 63.5 & 61.6 & 60.2 & 9.03 & \underline{50.6} & \underline{80.0} & 72.0 & 46.1 & 48.6 & 63.6 & 61.4 & 60.3 \\
 & KL Div & \underline{8.91} & 49.1 & 79.8 & 72.0 & 45.8 & 50.2 & 63.2 & 61.3 & 60.2 & 9.02 & 49.7 & 79.4 & 72.5 & 46.2 & 49.1 & \underline{63.9} & \underline{62.1} & 60.4 \\
 & HAWQ-v2 & \underline{8.91} & \underline{49.7} & \textbf{80.3} & 72.5 & \underline{45.9} & 50.1 & \underline{64.1} & \underline{63.3} & \underline{60.8} & \underline{9.00} & 49.7 & \underline{80.0} & \underline{73.4} & \underline{46.7} & \underline{49.2} & 63.6 & \underline{62.1} & 60.7 \\
 & Fisher & 9.28 & 48.3 & \underline{80.2} & \underline{72.9} & 45.6 & \underline{50.5} & 63.5 & 62.0 & 60.4 & 9.16 & 48.9 & 79.9 & 70.6 & 46.4 & 48.4 & 62.7 & 59.9 & 59.5 \\
 & CoopQ & 9.89 & 44.6 & 78.7 & 70.5 & 43.4 & 46.4 & 59.8 & 57.8 & 57.3 & 10.36 & 45.6 & 78.8 & 69.6 & 39.5 & 41.3 & 51.7 & 50.5 & 53.9 \\
 & LIM & 10.19 & 46.8 & 79.2 & 71.3 & 44.5 & 48.5 & 61.1 & 60.0 & 58.8 & 10.32 & 46.1 & 78.7 & 71.0 & 42.2 & 44.9 & 58.9 & 57.5 & 57.0 \\
\rowcolor{gray!15} & \textbf{\method{}} & \textbf{8.37} & \textbf{52.7} & 79.9 & \textbf{73.6} & \textbf{49.2} & \textbf{52.7} & \textbf{69.1} & \textbf{66.1} & \textbf{63.3} & \textbf{8.37} & \textbf{52.1} & \textbf{80.9} & \textbf{73.5} & \textbf{49.3} & \textbf{53.0} & \textbf{69.1} & \textbf{66.5} & \textbf{63.5} \\ \cline{1-20}
\multirow{7}{*}{4.0} & Quant. Err. & 8.71 & 49.5 & \underline{80.5} & \textbf{73.2} & 46.9 & \underline{51.5} & 65.1 & \underline{64.2} & \underline{61.6} & 8.76 & \underline{50.5} & 80.3 & \textbf{74.0} & 47.7 & 49.6 & 65.3 & \underline{63.8} & 61.6 \\
 & KL Div & 8.76 & 50.3 & \textbf{80.7} & 73.0 & \underline{47.1} & 51.2 & 64.7 & 63.5 & 61.5 & 8.81 & 50.2 & 80.0 & 72.7 & 47.4 & 49.2 & 65.0 & 63.3 & 61.1 \\
 & HAWQ-v2 & \underline{8.58} & \underline{50.5} & 80.3 & 72.3 & 46.7 & 51.1 & \underline{65.8} & 63.5 & 61.5 & \underline{8.66} & 49.6 & \textbf{80.5} & 72.8 & \underline{48.2} & 49.5 & 64.8 & 62.8 & 61.2 \\
 & Fisher & 8.82 & 49.1 & \underline{80.5} & 72.7 & 46.1 & 51.4 & 65.0 & 63.8 & 61.2 & 8.90 & 49.4 & 80.1 & \underline{73.2} & 46.5 & 49.2 & 65.2 & 62.9 & 60.9 \\
 & CoopQ & 9.48 & 45.3 & 79.1 & 71.7 & 43.6 & 47.5 & 61.8 & 59.4 & 58.3 & 9.94 & 46.1 & 78.5 & 69.1 & 40.9 & 43.1 & 53.8 & 52.2 & 54.8 \\
 & LIM & 9.80 & 48.1 & 79.7 & 72.8 & 45.2 & 50.9 & 64.7 & 62.9 & 60.6 & 9.94 & 49.0 & 79.3 & 72.9 & 47.4 & \underline{50.6} & \underline{65.5} & 62.0 & 61.0 \\
\rowcolor{gray!15} & \textbf{\method{}} & \textbf{8.20} & \textbf{53.4} & 80.0 & \underline{73.1} & \textbf{49.3} & \textbf{53.2} & \textbf{69.8} & \textbf{67.0} & \textbf{63.7} & \textbf{8.23} & \textbf{53.8} & \underline{80.4} & \textbf{74.0} & \textbf{50.0} & \textbf{52.9} & \textbf{69.7} & \textbf{67.2} & \textbf{64.0} \\ \cline{1-20}
\hline
\end{tabular}%
}
\end{table}

The margins are largest where quantization is most aggressive, and this is
where a context-aware score should matter most: at tight budgets many
modules sit at the lowest bitwidths, upstream contexts drift far from full
precision, and the resulting errors are large, irreversible, and propagate
downstream, so misallocating even a few modules is costly. On Llama-3.2-3B
at the 1.25\,GB budget, the tightest setting we evaluate, \method{} improves
average accuracy over the best baseline by 7.1 points under AWQ (55.6 vs.\
48.5) and 8.0 points under GPTQ (54.0 vs.\ 46.0), while reducing perplexity
from 12.43 to 10.70 and from 13.35 to 11.19, respectively. The gains are not
confined to a single task: at this budget \method{} improves WinoGrande by
roughly 5 points and the MMLU categories by up to 11 points over the
strongest baseline in each column. The pattern persists at 1.5\,GB, where
the GPTQ margin is 9.1 points. On Llama-2-7B at 3.0\,GB the margin is 4.2
points under AWQ, and on Mistral-7B the improvement ranges from 2.1 to 3.7
points across budgets and techniques.

\begin{figure}[t]
    \centering
    \includegraphics[width=\textwidth]{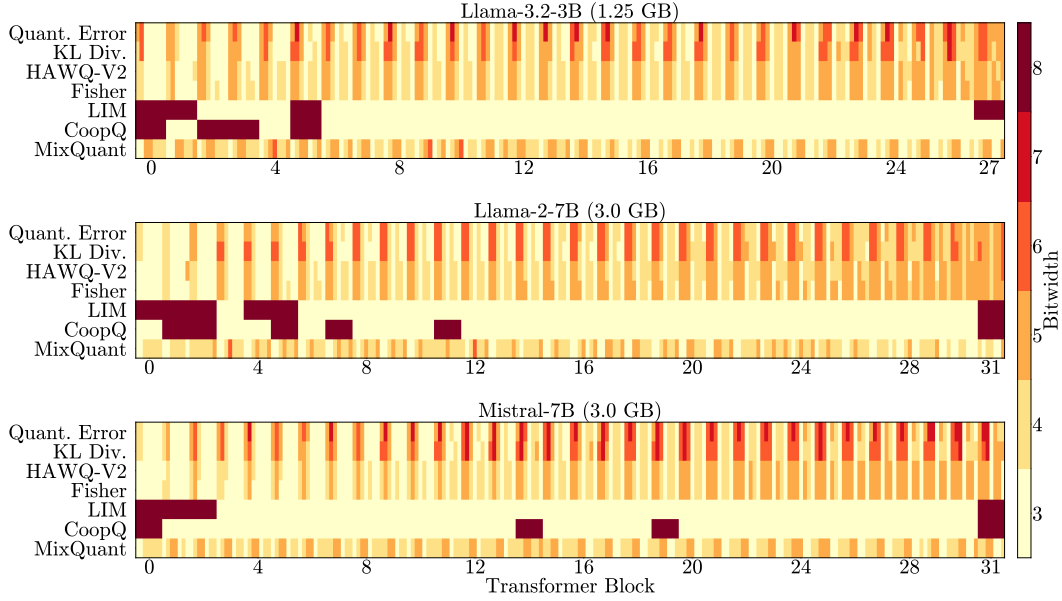}
    \caption{Per-linear-layer bitwidth allocations by method.
    For Llama-3.2-3B (1.25\,GB), Llama-2-7B (3.0\,GB), and Mistral-7B (3.0\,GB),
    the $x$-axis runs over linear layers, labelled by transformer block;
    the rows are scoring methods; shade encodes the bitwidth assigned to each layer.
    LIM and CoopQ collapse to the extremes of $\mathcal{B}$, while the others spread across the middle bitwidths.}
    \label{fig:baseline-bit-histograms}
\end{figure}

Notably, LIM and CoopQ are not the
strongest baselines in most cells. The FP16-prior metrics paired with our
ILP allocator frequently outperform them. Both LIM and CoopQ were designed and
validated on small bitwidth sets, typically $\{2, 3, 4\}$, where their
tendency to push allocations toward the extremes of the set is harmless
because the extremes are never far apart. Our setting expands the choice set
to $\mathcal{B} = \{3, \dots, 8\}$, which enlarges the plan space by orders
of magnitude and widens the gap between extremes.
Figure~\ref{fig:baseline-bit-histograms} shows the bitwidth distributions
of the resulting plans: LIM and CoopQ concentrate mass at the extremes of
$\mathcal{B}$, while \method{} spreads allocations across the middle
bitwidths.

On the Llama models, the margins narrow as the budget approaches the memory
of the highest uniform bitwidth: on Llama-2-7B the AWQ improvement falls
from 4.2 points at 3.0\,GB to 1.5 points at 4.0\,GB, and under GPTQ to 0.8.
This is expected. A generous budget places most modules at high bitwidths
under any reasonable allocation, and high-bitwidth quantization is robust
enough that the remaining decisions carry little weight, so accuracies
converge toward the plateau of the underlying technique. The value of
context-aware scoring is concentrated in the low-budget regime.

\paragraph{Component ablation.}
Table~\ref{tab:mixquant-ablation} isolates the contribution of the allocator on
Llama-3.2-3B at the 1.25\,GB budget. \emph{Mem.\ budget} removes the
distortion scores entirely: bits are assigned by an ILP whose only objective
is to fill the budget, i.e., $\min\,(M - \mathrm{mem}(\mathbf{b}))$ subject
to $\mathrm{mem}(\mathbf{b}) \le M$, so the allocation reflects module sizes
alone. \emph{Greedy (rev.)} runs our greedy allocator with the candidate
ordering reversed, applying the \emph{worst}-scoring upgrade first.
\emph{Stage~1 + Stage~2 + ILP} keeps the full scoring pipeline but replaces
the greedy solver with an ILP solution of \eqref{eq:mckp}. \method{}
is the full method: the efficiency-greedy allocator with the tail penalty of
Section~\ref{sec:tail-regularization}.

\begin{table}[t]
\centering
\caption{Component ablation on Llama-3.2-3B at the 1.25\,GB budget. 
Accuracies (\%); Wiki.\ = word perplexity ($\downarrow$). Avg excludes Wiki.}
\label{tab:mixquant-ablation}
\renewcommand{\arraystretch}{1.5}
\resizebox{\textwidth}{!}{%
\begin{tabular}{c|l|c|lllllllll|lllllllll}
\hline
\rowcolor[HTML]{FFFFFF} 
\cellcolor[HTML]{FFFFFF} & \cellcolor[HTML]{FFFFFF} & \cellcolor[HTML]{FFFFFF} & \multicolumn{9}{c|}{\cellcolor[HTML]{FFFFFF}AWQ} & \multicolumn{9}{c}{\cellcolor[HTML]{FFFFFF}GPTQ} \\ \cline{4-21} 
\multirow{-2}{*}{\cellcolor[HTML]{FFFFFF}Model} & \multirow{-2}{*}{\cellcolor[HTML]{FFFFFF}\begin{tabular}[c]{@{}l@{}}Mem \\ (GB)\end{tabular}} & \multirow{-2}{*}{\cellcolor[HTML]{FFFFFF}Variant} & Wiki. & Arc & PIQA & Wino & STEM & Hum. & Soc. & \multicolumn{1}{l|}{Othr} & Avg & Wiki. & Arc & PIQA & Wino & STEM & Hum. & Soc. & \multicolumn{1}{l|}{Othr} & Avg \\ \hline
\rowcolor[HTML]{FFFFFF} 
\cellcolor[HTML]{FFFFFF} & \cellcolor[HTML]{FFFFFF} & Mem. budget & 12.97 & 35.8 & 73.1 & \underline{62.2} & 37.7 & 37.3 & 47.6 & \multicolumn{1}{l|}{\cellcolor[HTML]{FFFFFF}49.3} & 49.0 & 16.19 & 34.4 & 71.8 & 60.2 & 32.0 & 32.0 & 36.7 & \multicolumn{1}{l|}{\cellcolor[HTML]{FFFFFF}38.3} & 43.6 \\
\rowcolor[HTML]{FFFFFF} 
\cellcolor[HTML]{FFFFFF} & \cellcolor[HTML]{FFFFFF} & Greedy (rev.) & 17.59 & 32.5 & 69.9 & 59.8 & 27.0 & 29.3 & 30.6 & \multicolumn{1}{l|}{\cellcolor[HTML]{FFFFFF}33.3} & 40.3 & 24.26 & 28.9 & 64.3 & 55.2 & 26.6 & 27.8 & 28.7 & \multicolumn{1}{l|}{\cellcolor[HTML]{FFFFFF}28.0} & 37.1 \\
\rowcolor[HTML]{FFFFFF} 
\cellcolor[HTML]{FFFFFF} & \cellcolor[HTML]{FFFFFF} & Stage 1 + Stage 2 + ILP & \underline{11.02} & \underline{41.8} & \underline{75.7} & \textbf{68.4} & \textbf{40.7} & \textbf{45.4} & \textbf{58.2} & \multicolumn{1}{l|}{\cellcolor[HTML]{FFFFFF}\textbf{59.0}} & \textbf{55.6} & \underline{11.33} & \textbf{42.5} & \underline{75.8} & \textbf{67.0} & \textbf{42.8} & \textbf{45.1} & \textbf{57.3} & \multicolumn{1}{l|}{\cellcolor[HTML]{FFFFFF}\textbf{57.0}} & \textbf{55.3} \\ \cline{3-21}
\rowcolor[HTML]{EFEFEF} 
\cellcolor[HTML]{FFFFFF}\multirow{-4}{*}[-22pt]{Llama-3.2-3B} & \multirow{-4}{*}{\cellcolor[HTML]{FFFFFF}1.25} & \textbf{MixQuant} & \textbf{10.75} & \textbf{43.9} & \textbf{75.8} & \textbf{68.4} & \underline{40.4} & \underline{43.2} & \underline{57.9} & \multicolumn{1}{l|}{\cellcolor[HTML]{EFEFEF}\underline{56.7}} & \underline{55.2} & \textbf{11.25} & \underline{42.2} & \textbf{76.2} & \underline{66.6} & \underline{39.8} & \underline{43.0} & \underline{53.7} & \multicolumn{1}{l|}{\cellcolor[HTML]{EFEFEF}\underline{55.6}} & \underline{53.9} \\ \hline
\end{tabular}%
}
\end{table}

The score-free and reversed variants collapse, losing 6--17 average points
against the full method, which confirms that the decoupled distortion table
carries real signal: allocating without it, or against it, is severely
punished at this budget. The comparison with the ILP variant addresses the
solver. The ILP is the natural upper bound for our surrogate objective, yet
\method{} matches it on average (55.2 vs.\ 55.6 under AWQ, 53.9 vs.\ 55.3
under GPTQ) and surpasses it on several individual tasks, including
perplexity under both techniques (10.75 vs.\ 11.02 and 11.25 vs.\ 11.33),
PIQA, and WinoGrande under GPTQ. The greedy solver is also the only variant
compatible with the adaptive setting: the ILP took upwards of two hours to
converge on some budgets, whereas the greedy solve costs
$O(L\,|\mathcal{B}| \log L)$ and runs in negligible time at deployment,
where a new budget must be served on arrival.

\begin{figure}[H]
    \centering
    \includegraphics[width=0.85\textwidth]{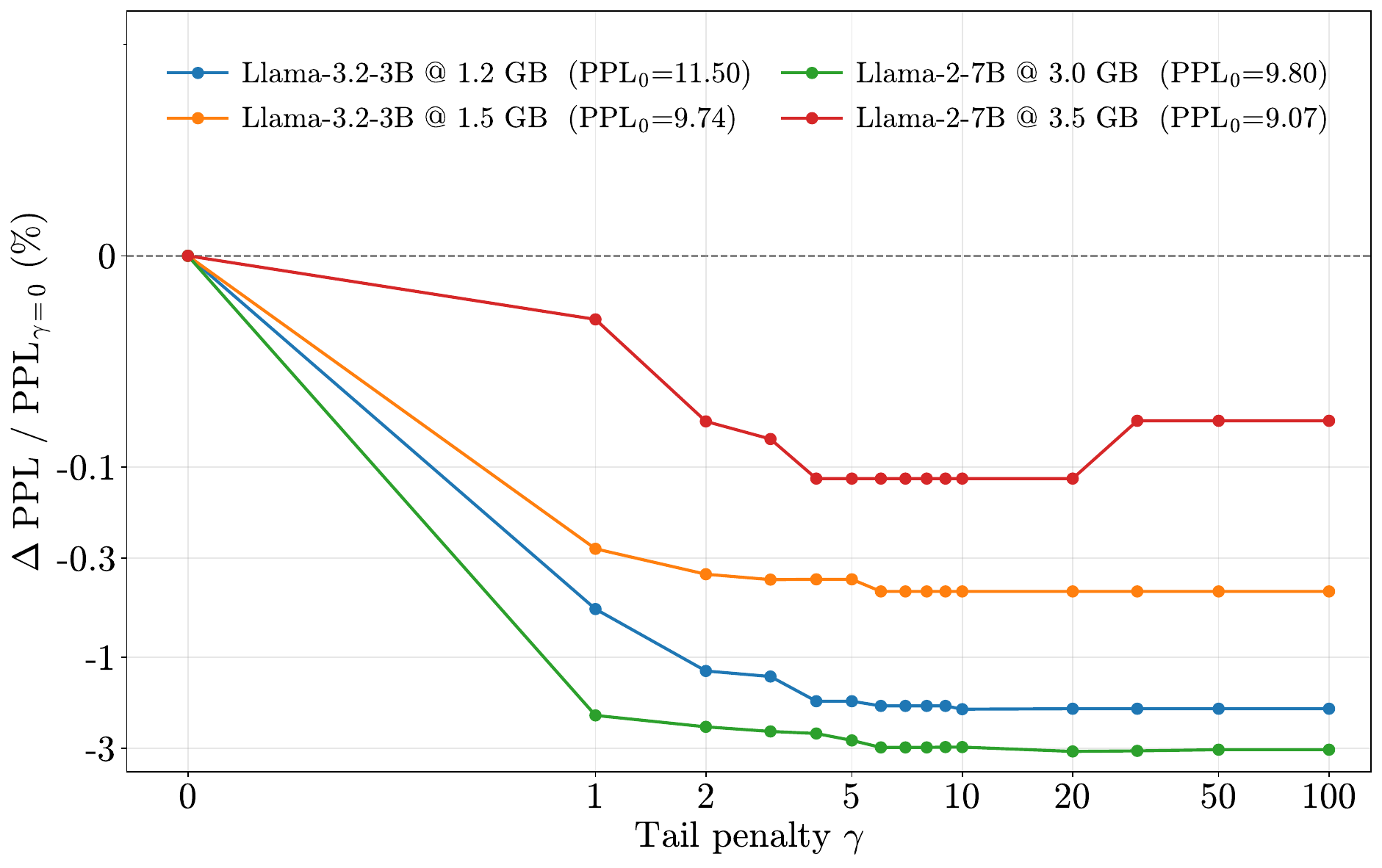} 
    \caption{Effect of the tail penalty. Relative change in WikiText-2 perplexity
versus the unregularized allocator ($\gamma=0$) as the tail-penalty strength
$\gamma$ increases, across four (model, budget) settings; $\mathrm{PPL}_0$ is the
$\gamma=0$ perplexity for each curve. (Section~\ref{sec:tail-regularization})}
    \label{fig:tail-sweep}
\end{figure}

\paragraph{Sensitivity to the tail penalty.}
Figure~\ref{fig:tail-sweep} sweeps the tail-penalty strength $\gamma$ and
reports the relative change in perplexity against the unregularized
allocator ($\gamma = 0$) across four (model, budget) settings. Two
observations follow. First, we observe that the penalty helps: no setting is
hurt relative to $\gamma = 0$, and perplexity typically decreases as $\gamma$
grows from 0.
Second, the effect saturates: beyond $\gamma \approx 5$ the
curves are flat. 
Saturation is expected from the mechanism of the penalty.
Once $\gamma$ is large enough to redirect the budget away from the
tail-most upgrades, only a small number of contested transitions remain
whose ordering the penalty can still change; increasing $\gamma$ further
re-ranks nothing, and the allocator returns the same plan. 

\FloatBarrier

\section{Conclusion}
\label{sec:mixquant-conclusion}

We presented \method{}, a technique-agnostic adaptive post-training quantization framework
in which one offline calibration serves any deployment budget
through a single inexpensive greedy solve.
It starts from a mismatch that prior adaptive quantizers overlook:
per-layer sensitivity measured against an FP16 network does not reflect
the fully quantized model that is actually deployed.
\method{} closes this gap with two components:
mean-field decoupled distortion scores, which marginalize each layer's error
over random quantized upstream contexts;
and plan-aware technique parameters, which are calibrated on
the activations the deployed plan induces rather than on FP16 activations.
It additionally penalizes retaining low-bit assignments through a tail regularizer
that steers spare budget away from layers still stuck at the lowest bitwidths.
Across Llama-3.2-3B, Llama-2-7B, and Mistral-7B, 
under both AWQ and GPTQ, \method{} produces consistently better
bit allocations than prior adaptive and mixed-precision baselines.
Its greedy allocator matches an ILP solver at negligible deployment cost.
Future work includes extending \method{} to jointly allocate precision across
weights, activations, and KV-cache, toward maximizing efficiency in adaptive
settings on the edge.

\bibliographystyle{plainnat}
\bibliography{references}

\end{document}